\documentclass[sigconf]{acmart}
\usepackage{booktabs} % For formal tables

\begin{document}
\title{Text Coherence Analysis Based on Deep Neural Network}
%\titlenote{Produces the permission block, and
%  copyright information}
%\subtitle{Extended Abstract}
%\subtitlenote{The full version of the author's guide is available as
%  \texttt{acmart.pdf} document}
\copyrightyear{2017} 
\acmYear{2017} 
\setcopyright{acmcopyright}
\acmConference{CIKM'17 }{November 6--10, 2017}{Singapore, Singapore}\acmPrice{15.00}\acmDOI{10.1145/3132847.3133047}
\acmISBN{978-1-4503-4918-5/17/11}

\fancyhead{}
\settopmatter{printacmref=false, printfolios=false}

\author{Baiyun Cui, Yingming Li\footnotemark[2], Yaqing Zhang and Zhongfei Zhang}
\affiliation{\institution{Zhejiang University, China}}
\email{baiyunc@yahoo.com, {yingming, yaqing, zhongfei}@zju.edu.cn}

% The default list of authors is too long for headers}
%\renewcommand{\shortauthors}{B. Trovato et al.}

\begin{abstract}
%In this paper, we introduce a novel coherence model using deep learning architecture to analyze the level of coherence in text. This model applies convolutional neural networks for learning an optimal distributional representation of input sentence, which can not only succeed in capturing semantic and syntactic information in sentence but also obtain the benefit of requiring no manual feature engineering and no external knowledge sources (parser trees). More importantly, we emphasize the role of interactions between sentences by injecting relational information into final representation to learn better coherence structure of the text. We test the proposed model on a widely used ordering task. The experiment results demonstrate that our approach achieves a significantly higher accuracy in distinguishing coherent and incoherent news articles from two domains and generates the state-of-the-art performance by an average 5.3\% absolute improvement over previous work.
In this paper, we propose a novel deep coherence model (DCM) using a convolutional neural network architecture to capture the text coherence. The text coherence problem is investigated with a new perspective of learning sentence distributional representation and text coherence modeling simultaneously. In particular, the model captures the interactions between sentences by computing the similarities of their distributional representations. Further, it can be easily trained in an end-to-end fashion. The proposed model is evaluated on a standard Sentence Ordering task. The experimental results demonstrate its effectiveness and promise in coherence assessment showing a significant improvement over the state-of-the-art by a wide margin.
\end{abstract}
%
% The code below should be generated by the tool at
% http://dl.acm.org/ccs.cfm
% Please copy and paste the code instead of the example below. 




\maketitle
\footnotetext[2]{Corresponding author}
\section{Introduction}
Coherence is a key property of any well-organized text. %This paper focus on coherence analysis, a central concern in natural language processing. 
It evaluates the degree of logical consistency for text and can help document a set of sentences into a logically consistent order, which is at the core of many text-synthesis tasks such as text generation and multi-document summarization. An example is shown in Table~\ref{tab:three} for the coherence problem.

Although coherence is significant in constructing a meaningful and logical multi-sentence text, it is difficult to capture and measure as the concept of coherence is too abstract. %Hence, aim for coherence assessment, our main effort lies in two key aspects, First, learning good representations for each sentence in  text to capture both lexical and syntactic information; Second, based on the representations we obtained, exploring intrinsic semantic and logical relationships between these sentences which reflect text structure, thus is able to improve accuracy in coherence analysis.
The problem of coherence assessment was first proposed in 1980s, and since then a variety of coherence analysis methods have been developed, such as the centering theory \cite{DBLP:journals/coling/PoesioSEH04,grosz1995centering} which establishes constraints on the distribution of discourse entities in coherent text, and the content approaches \cite{DBLP:conf/naacl/BarzilayL04,fung2006one} as the extensions of HMMs for global coherence which consider text as a sequence of topics and represent topic shifts within a specific domain.% each characterized by a particular distribution of lexical items.
%\begin{figure}
%%\setlength{\abovecaptionskip}{10pt} 
%\setlength{\belowcaptionskip}{-15pt}
%  \centering
%    \includegraphics[width=0.48\textwidth]{three}
%  \caption{Examples of a coherent text and an incoherent one.}
%  \label{fig:three}
%\end{figure}

%\begin{table}[]
%\renewcommand\arraystretch{1.2}  
%\centering
%\caption{Examples of a coherent text and an incoherent one.}
%%\vspace*{2.0em}
%\label{tab:three}
% \scalebox{0.85}{
%\begin{tabular}{|cll|}
% \hline
%\multicolumn{1}{|}{} & \multicolumn{1}{c}{\textbf{Text 1}} & \multicolumn{1}{c|}{\textbf{Text 2}} \\ \hline
%1                      & Tom loves reading books.             & Tom loves reading books.             \\
%2                      & He prefers reading books at library. & He missed his lunch today.           \\ 
%3                      & So he always goes to library.        & So he always goes to library.        \\ 
%label                  &\multicolumn{1}{c}{label=1 (coherent)}                         &\multicolumn{1}{c}{label=0 (incoherent)}                       \\ \hline
%\end{tabular}}
%\end{table}

\begin{table}[]
\renewcommand\arraystretch{1.1}  
\centering
\caption{Examples of a coherent text and an incoherent one.}
%\vspace*{-1em}
\label{tab:three}
\scalebox{0.93}{
\begin{tabular}{|l|l|}
\hline
\multicolumn{1}{|c|}{\textbf{Text 1}}             & \multicolumn{1}{c|}{\textbf{Text 2}}               \\ \hline
Tom loves reading books.                          & Tom loves reading books.                           \\ 
He prefers reading books at library.              & He missed his lunch today.                         \\ 
So he always goes to library.                     & So he always goes to library.                      \\ \hline
\multicolumn{1}{|c|}{label=1\; \; (coherent) }& \multicolumn{1}{c|}{label=0 \; \;(incoherent)}\\ \hline
\end{tabular}}
%\vspace*{-1.8em}
\end{table}

Another widely used type of approaches in the literature is to encode input text into a set of sophisticated lexical and syntactic features, and then apply machine learning methods (e.g., SVM) to measure coherence between these representations based on the features. Features being explored include entity-based features \cite{DBLP:journals/coling/BarzilayL08},  syntactic patterns \cite{DBLP:conf/emnlp/LouisN12}, conference clues to ordering \cite{DBLP:conf/acl/ElsnerC08a}, named-entity features \cite{DBLP:conf/acl/ElsnerC11a}, and others. But, identifying and defining those features are always an empirical process which requires considerable experience and domain expertise. %Moreover, it can be computationally expensive and needs a large number of knowledge bases, such as syntactic parsers, lexicons and other external tools. Additionally, adapting to different texts in new domains calls for additional efforts to tune feature extraction algorithms. On the contrary, our approach do not depend on these complex manual feature extraction algorithms and efficiently obtain better performance in the results.

Recently, a promising coherence framework \cite{DBLP:conf/emnlp/LiH14a} has been proposed based on a deep learning framework, where it adopts recurrent and recursive neural networks  \cite{DBLP:conf/interspeech/MikolovKBCK10,DBLP:conf/interspeech/MesnilHDB13} in computing vectors for input sentences. %In particular, The recurrent sentence representation views input as a sequence of tokens and captures information from both current token and the past, regarding sentential compositionality at each step. Besides, the recursive networks rely on the structure of parse trees and use children nodes to compute a representation for its parent node of the tree. Both architectures have been successfully applied in many NLP tasks, including language modeling \cite{DBLP:conf/interspeech/MikolovKBCK10} and spoken language understanding \cite{DBLP:conf/interspeech/MesnilHDB13}. 
However, it pays little attention to essential semantic interactions between sentences, which are also necessary in coherence assessment. Furthermore, in the recurrent framework, terms are simply piled up within a sentence such that long-distance dependencies are difficult to capture due to the vanishing gradient problem \cite{DBLP:journals/tnn/BengioSF94} and on the other hand, the recursive neural network still suffers from a severe dependence on external resources to construct its syntactic trees.

%It has been shown that coherence assessment task can be efficiently tackled with distributional word embeddings by deep learning framework. The method of distributed representations  for words was first proposed in \cite{rumelhart1988learning} and one of  the main advantages is to alleviate the problem of data sparseness. Such embeddings can be derived from several approaches, e.g., by counting the co-occurring frequencies of  words in a large corpora or adopting some neural language models. Recently, deep learning architectures have been successfully applied to various natural language processing tasks, including sentence classification \cite{DBLP:conf/acl/KalchbrennerGB14} and  modelling text pairs \cite{DBLP:conf/nips/LuL13}. It also has been explored to learn word embeddings in an unsupervised manner for advanced sentence representations. Deep connectionist architectures are able to effectively capture the meaning of individual words to a continuous representation of sentence. In particular, it has been widely known that convolutional neural networks have a good performance in learning to embed dense sentence representations into low-dimensional vector space in a hierarchical way, while preserving syntactic and semantic aspects of the original input at higher level, which leads to state-of-art results in many NLP tasks \cite{DBLP:conf/acl/KalchbrennerGB14,DBLP:journals/corr/YuHBP14}. Then, with these optimal sentence representations, neural networks can to some extent learn coherent sentence structure by capturing interesting semantic relationships between them.

To overcome the above limitations, in this work, we present a novel deep coherence model (DCM) based on convolutional neural networks to learn coherence for the given text. We study the text coherence problem with a new perspective of learning sentence distributional representation and text coherence modeling simultaneously. In particular, word embeddings are first explored to generate sentence matrix for each sentence \cite{DBLP:conf/sigir/SeverynM15,DBLP:journals/corr/SeverynM16,DBLP:conf/emnlp/Kim14,DBLP:conf/nips/MikolovSCCD13}, and then sentence models map sentences to distributional vectors in parallel, which are used for learning coherence between them. Further, interactions between sentences are captured by computing the similarities of their distributional representations. Finally, the sentence vectors and their corresponding similarity scores are concatenated together to estimate the text coherence.

Our work differs from the existing approaches in several important aspects: 1) we propose a distributional sentence model based on convolutional neural networks (CNNs) to map input sentences to advanced representations; 2) our architecture uses intermediate sentence vectors to compute their similarity scores and includes them in the final representation, which constitutes a much richer representation of text.%; 4) this model automatically learns the properties of coherent texts from two large corpus in an low-cost way, without any predefined knowledge base, thus obviating the need for manual feature engineering and external resources; 5) we train deep neural networks in an end-to-end fashion starting from the input text to a final coherence probability score, which is pretty flexible to be used in new tasks and domains.

The proposed model is evaluated on a standard Sentence Ordering task. The experimental results demonstrate the effectiveness and promise in coherence assessment showing considerable improvements over the state-of-the-art literature \cite{DBLP:conf/emnlp/LiH14a} by a wide margin.

%In the following section, we show a brief overview of existing work on coherence analysis. Then, introduce distributed sentence representation and describe how to evaluate coherence using our deep learning framework. Next, present data and experimental results, followed by a conclusion in the end.

%\section{Related Work}

\section{Model Construction}
\label{sec:3}

In this section, we first introduce how to compute distributional sentence vectors based on  CNNs and word embeddings. Then, a framework for evaluating the coherence of a sequence of sentences is proposed with the sentence representations.

\subsection{Distributional Sentence Representation}
\begin{figure}
\setlength{\belowcaptionskip}{-5pt}
  \centering
    \includegraphics[width=0.5\textwidth]{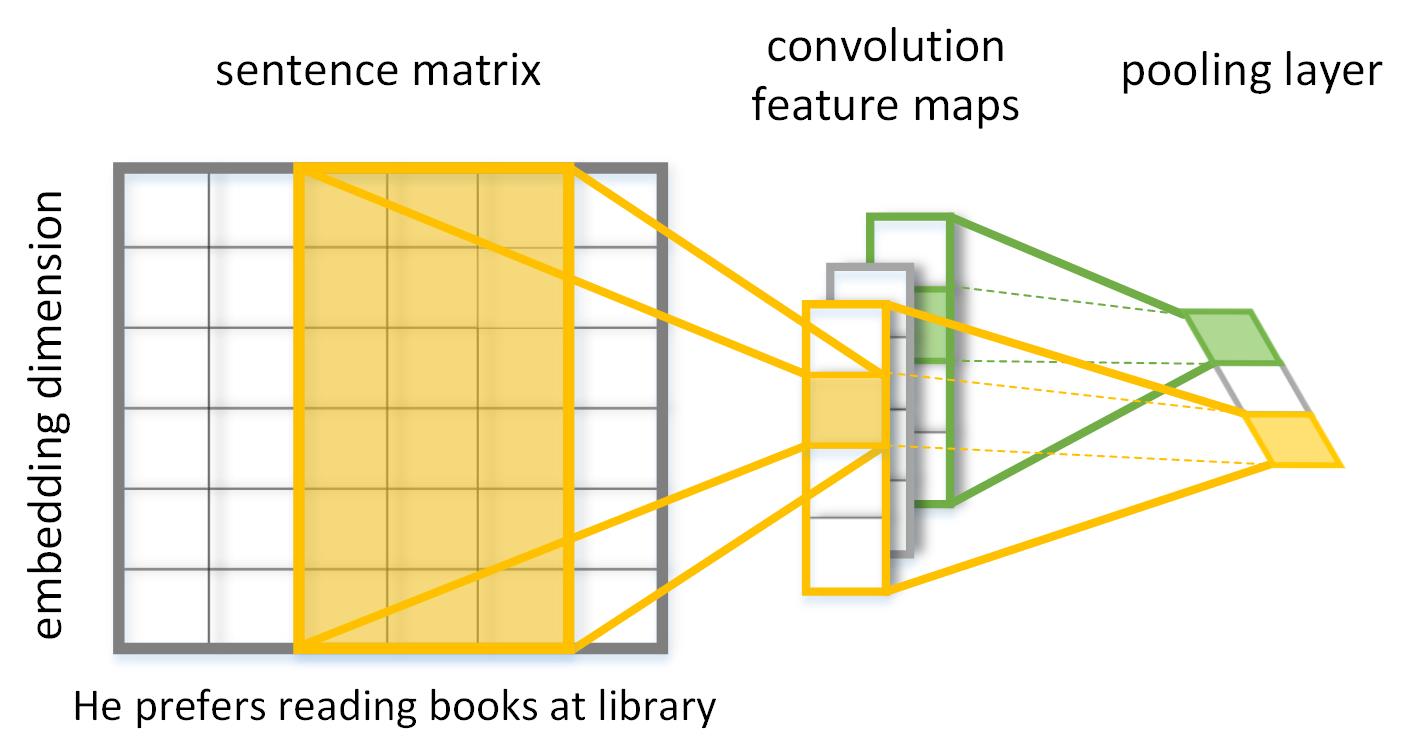}
%\vspace{-2em}
  \caption{The sentence model based on a CNN for distributional representation.}
 \label{fig:small}
%\vspace{-1em}
\end{figure}

Given a sequence of sentences, as is shown in Figure~\ref{fig:small}, the proposed sentence model is able to map each sentence into a distributional vector, % where word embeddings are first augmented with additional dimensions to encode overlapping information around this sentence, 
and then the dense sentence representation is transformed through a wide convolutional layer, a non-linearity and a max pooling layer into a low-dimensional and real-valued feature vector.

In the following, we describe the main building blocks of our sentence model in details: sentence matrix and CNN including convolutional layers, activations and pooling layers.

%\vspace{-4em}
\subsubsection{Sentence Matrix}
Since the input sentence is comprised of several raw words which cannot be directly processed by subsequent layers of the network, we adopt distributional word embeddings to translate the words into real-valued feature vectors and then combine them to form a sentence matrix.% which can not only sufficiently represent this sentence but also well suited with ConvNets and later computation.

The input sentence $\mathbf{s}$ consists of a sequence of words: $[w_1,\dots,w_{|s|}]$, where $|s|$ denotes the total number of words within the sentence. Word embeddings matrix  $\mathbf{W} \in \mathbb{R}^{ d\times|V|}$ is formed by concatenating embeddings of all words in a finitely sized vocabulary $V$, where $d$ denotes the dimension of this embedding. Each word is mapped to integer indices $1,\dots ,|V|$ in vocabulary $V$ and then represented by a distributional vector $\mathbf{w} \in \mathbb{R}^ {d}$ looked up in this word embeddings matrix. Hence, a sentence matrix $\mathbf{S} \in \mathbb{R}^{ d\times|s|}$ is established for each input sentence $\mathbf{s}$, where each $i$-th column represents a word embedding $\mathbf{w}_i$ of the $i$-th word in the sentence: 
%{\setlength\abovedisplayskip{1pt}
%\setlength\belowdisplayskip{1pt}
\begin{equation}
   \mathbf{S} = [\mathbf{w}_1, \mathbf{w}_2,\dots, \mathbf{w}_{|s|}] 
\end{equation}

So far we have obtained a sentence matrix $\mathbf{S}$. In the following, a CNN is applied to the input sentence matrix to capture higher level semantic concepts, using a series of transformations including convolution, nonlinearity and pooling operations.

\subsubsection{Convolutional Neural Network}
%The main section in sentence representation are three distributional models based on ConvNets. These underlying models map input sentences to intermediate representations which can then be used for learning the semantic coherence among them.

%\textbf{Convolutional feature maps\\}
The aim of the convolutional layer is to extract useful patterns using different filters which represent a variety of significant features of the input sentence. 

Corresponding to the input $\mathbf{S} \in \mathbb{R}^{d\times|s|}$, a convolution filter is also a matrix of weights $\mathbf{F} \in \mathbb{R}^{d\times m}$ with width $m$ and has the same dimensionality $d$ as the given sentence matrix. As shown in Figure~\ref{fig:small}, the filter slides along the column dimension of $\mathbf{S}$ producing a vector $\mathbf{c} \in \mathbb{R}^{|s|-m+1}$ as an output, where each component is computed as follows: %$ c_i=(\mathbf{S} \ast \mathbf{F})_i=\sum_{k,j}(\mathbf{S}_{[:,i-m+1:i]}\otimes \mathbf{F})_{kj}$,
%\[\mathbf{c}_i=(\mathbf{s} \ast \mathbf{f})_i=\mathbf{s}_{[i-m+1:i]}^{T} \cdot \mathbf{f}\] %= \sum_{k=i}^{i+  m-1} s_kf_k\]
{\setlength\abovedisplayskip{1pt}
\setlength\belowdisplayskip{5pt}
\begin{equation}
   c_i=(\mathbf{S} \ast \mathbf{F})_i=\sum_{k,j}(\mathbf{S}_{[:,i-m+1:i]}\otimes \mathbf{F})_{kj}
\end{equation}}
where $\mathbf{S}_{[:,i-m+1:i]}$ is a matrix slice with size $m$ along the columns and $\otimes$ is the element-wise multiplication. Essentially, in order to capture more features and build a richer representation for the input sentence, the networks apply a set of filters sequentially convolved with the distributional sentence matrix $\mathbf{S}$. Such filter banks $\mathbf{F} \in \mathbb{R}^{n\times d\times m}$ work in parallel generating multiple feature maps of dimension $n\times(|s|-m+1)$. 

%\textbf{Activation units and pooling layer\\}
After convolution operations, we apply a non-linear activation function $\alpha()$ to learn nonlinear decision boundaries and adopt a rectified linear (ReLU) function defined as $max(0,\mathbf{x})$ which can not only speed up the training process but also sometimes increase the accuracy. Furthermore, we add $max$ pooling layer to the distributional sentence model aiming to reduce the representation and aggregate the information. This $max$ pooling operates on columns of the feature map matrix $\mathbf{C}$ and enables to return the maximum value of the output from the convolutional layer as follows: pool($\mathbf{c}$): $\mathbb{R}^{|s|-m+1} \rightarrow \mathbb{R}$, which has just passed through the activation function. 

%So far, we have already demonstrated the strategy used in our sentence model that a convolutional layer passed through the activation function together with pooling layer can act as a nonlinear feature extractor to learn better representations for a given sentence. In the following, we will present our deep learning architecture for computing text coherence score based on these distributional representations we have obtained at present.

\subsection{Coherence Computation}
\label{sec:2.2}
Here we explain how to map several input sentences to the final coherence probability and provide a full description of the remaining components in the networks, e.g.,  similarity matrix, join layer, hidden and softmax layer. 

\begin{figure}
  \centering
    \includegraphics[width=0.5\textwidth]{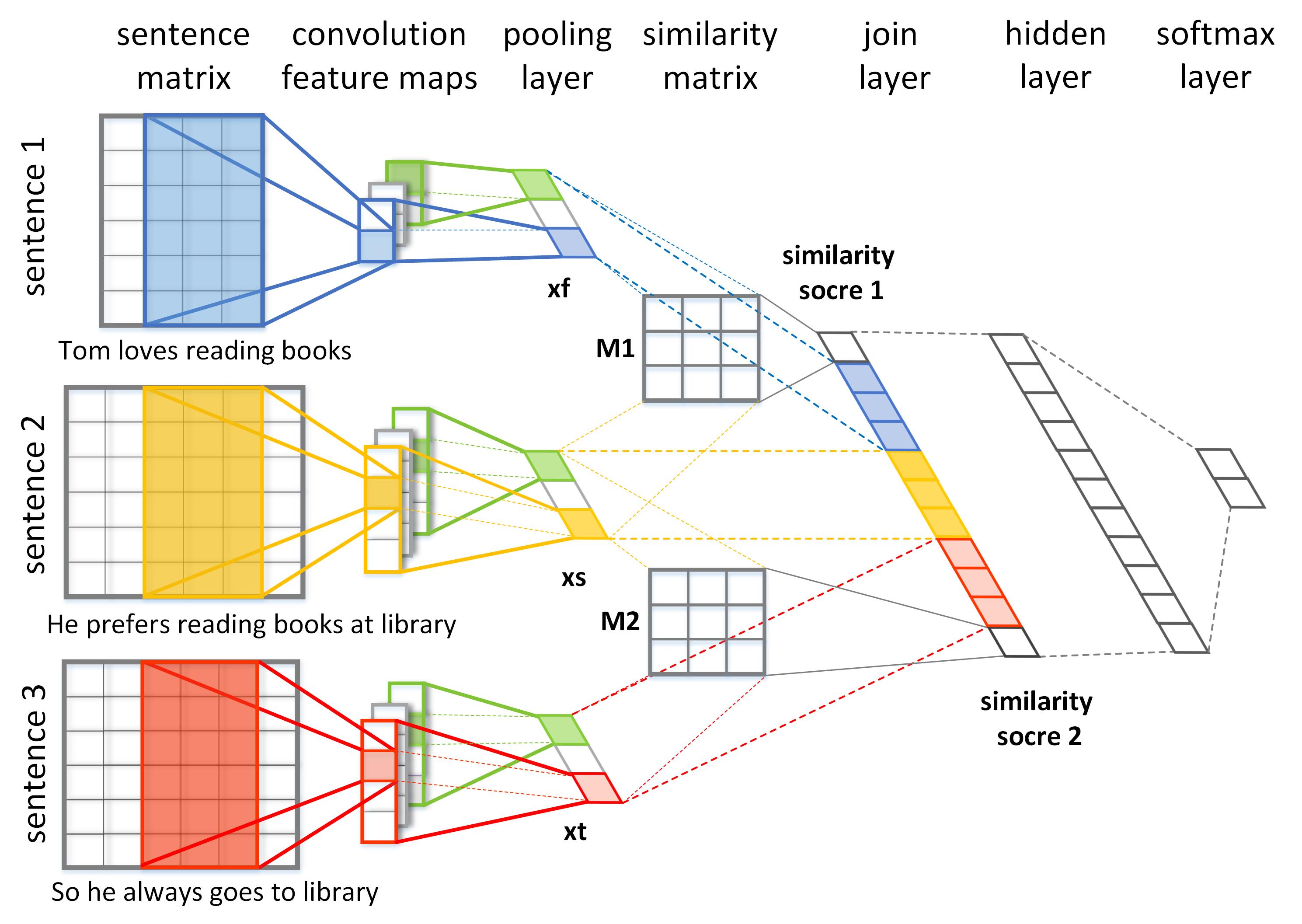}
%\vspace{-2.5em}
  \caption{The architecture of our deep coherence model (DCM) for text coherence analysis with two matrices to encode similarity between adjacent sentences.} % with additional dimensions to extract over-lapping information and two matrix to encode similarity between adjacent sentences}
\label{fig:big}
%\vspace{-1em}
\end{figure}

We first define a window of sentences as a clique $q$ and associate each clique with a label $y_q$ that indicates its coherence, where $y_q$ takes the value 1 if coherent, and 0 otherwise. Consequently, for a document $D$ consisting of $N$ sentences $D = \{s_1,s_2,\dots,s_N\},$ it is comprised of $N_d$ cliques. Taking window size $L=3$ for example, $N_d=N-2$, and the cliques we need to consider are as follows:
%{\setlength\abovedisplayskip{1pt}
%\setlength\belowdisplayskip{1pt}
\begin{equation}
< s_1, s_2, s_3 >, < s_2, s_3, s_4 >,\dots,< s_{N-2}, s_{N-1}, s_N >
\end{equation}To articulate clearly the coherence computation methodology, in the following we use the case of a clique of three neighboring sentences to present the methodology and the architecture of our model is shown in Figure~\ref{fig:big}. The method, however, can be implemented using a clique of any number of neighboring sentences and in fact in the experiments we have implemented and evaluated the method in different clique sizes. It appears that the performance differences for different clique sizes are not significant.\\ 
%\textbf{Similarity computation\\}
%This section introduce a great improvement we made on previous approaches \cite{DBLP:conf/emnlp/LiH14a} in the same task. 
\textbf{Similarity computation.} Since sentences in coherent text always talk about a main topic and share some events and emotions in common, we compute sentence-to-sentence semantic similarity to encode this essential information which can certainly produce positive effects on coherence assessment.
Assume that the output of our sentence model is three distributional representations: the first one $\mathbf{x}_f$, the second one $\mathbf{x}_s$ and the third one $\mathbf{x}_t$. Following the approach used in \cite{DBLP:conf/pkdd/BordesWU14}, we define the similarity between any neighboring sentence vectors as follows: 
%{\setlength\abovedisplayskip{1pt}
%\setlength\belowdisplayskip{1pt}
\begin{equation}
   Sim(fs) = \mathbf{x}_f^T\mathbf{M}_1\mathbf{x}_s
\end{equation}
the similarity matrix $\mathbf{M}_1$ is a parameter of the network that can be optimized during the training. In this model, more common elements between these two vectors, closer $\mathbf{x}'_s = \mathbf{M}_1\mathbf{x}_s$ is to $\mathbf{x}_f$, and thus the higher similarity score $Sim(fs)$.\\
%\textbf{Join layer\\}
\textbf{Join layer.} For the coherence assessment of the three input sentences, similarity computation produces two single scores: $ Sim(fs)$ and  $Sim(st)$ capturing syntactic and semantic aspects of the similarity from the input three-sentence text. Additionally, along with two similarity scores, our architecture also includes intermediate representations of the three sentences into the final vector
%{\setlength\abovedisplayskip{1pt}
%\setlength\belowdisplayskip{1pt}
\begin{equation}
 \mathbf{x}_{join} = [\mathbf{x}_f^T, Sim(fs),  \mathbf{x}_s^T,   Sim(st),  \mathbf{x}_t^T]
\end{equation}
which together constitute a much richer final representation for computing the final coherence probability.\\
%\textbf{Hidden layer\\}
%The concatenating vector is then passed through a fully connected hidden layer, this additional layer right before the softmax layer (described next) allows for modeling interactions between the components of the joined representation vector by computing
\textbf{Hidden layer.} The hidden layer performs the following transformation: $\mathbf{h}=f(\mathbf{W}_h\mathbf{x}_{join} + \mathbf{b}_h),$ where $\mathbf{W}_h$ is the weight matrix of the hidden layer, $\mathbf{b}_h$ is a bias vector and $f()$ is the non-linearity function. The output of the hidden layer $\mathbf{h}$ can be viewed as a final abstract representation obtained by a series of transformations from the input layer through a series of convolutional and pooling operations.\\
%\textbf{Softmax layer\\}
%Finally, given that we choose to model the coherence assessment task as a binary classification, 
\textbf{Softmax layer.} 
The output of the hidden layer $\mathbf{h}$ is further fed to the fully connected softmax classification layer, and the coherence probability of this three-sentence text can be summarized as: 
%{\setlength\abovedisplayskip{1pt}
%\setlength\belowdisplayskip{1pt}
%\begin{equation}
%\mathbf{v} = \text{softmax}(\mathbf{W}_s\mathbf{h} + \mathbf{b}_s)
%\end{equation}
\begin{equation}
p(y_q|\mathbf{x}_{join}) = \text{sigmod}(\mathbf{W}_s\mathbf{h} + \mathbf{b}_s)
\end{equation}
where $\mathbf{W}_s$ is a weight matrix and $\mathbf{b}_s$ denotes the bias. %And $p(y_c=0)$ can also be interpreted as the degree of the text is not coherent.

\subsection{Training}
Parameters in our deep neural network include: the word embeddings matrix $\mathbf{W}$% and word overlap feature matrix $\mathbf{W}_o$
, filter weights and biases of the convolutional layers in sentence model, two similarity matrices $\mathbf{M}_1$ and $\mathbf{M}_2$, and parameters of the hidden and softmax layers. We use $\theta$ to represent them: 
%{\setlength\abovedisplayskip{1pt}
%\setlength\belowdisplayskip{1pt}
\begin{equation}
\theta=\{ \mathbf{W}; \mathbf{F}_{x_f}; \mathbf{b}_{x_f};\mathbf{F}_{x_s}; \mathbf{b}_{x_s}; \mathbf{F}_{x_t}; \mathbf{b}_{x_t}; \mathbf{M}_1; \mathbf{M}_2; \mathbf{W}_h; \mathbf{b}_h; \mathbf{W}_s; \mathbf{b}_s\}
\end{equation}
%\begin{displaymath}
% \mathbf{b}_{x_t}; \mathbf{M}_1; \mathbf{M}_2; \mathbf{w}_h; \mathbf{b}_h; \mathbf{w}_s; \mathbf{b}_s\},
%\end{displaymath}
and the negative conditional log-likelihood of the training set is: 
{\setlength\abovedisplayskip{1pt}
\setlength\belowdisplayskip{1pt}
\begin{equation}
C=-\text{log}\prod^{N_t}_{i=1} p(y_q|\mathbf{g}^{(i)};\theta)
\end{equation}}where $\mathbf{g}^{(i)}=(\mathbf{x}_f^i,\mathbf{x}_s^i,\mathbf{x}_t^i)$ denotes the $i$-th training clique of three neighboring sentences and $N_t$ indicates the number of training cliques. We train the overall model to minimize this function and optimize parameters in the network by computing their gradients within shuffled mini-batches based on back propagation algorithm.

\section{Document Coherence Assessment}
\label{sec:4}
In this section, we apply the proposed framework to evaluate coherence for any documents with varying lengths. With the definition of a clique in Section~\ref{sec:2.2}, the function to compute the coherence score for a whole document is given by \cite{DBLP:conf/emnlp/LiH14a}: 
{\setlength\abovedisplayskip{1pt}
\setlength\belowdisplayskip{1pt}
\begin{equation}
 S_D=\prod_{q \in D} p(y_q=1)  
\end{equation}}
It is reasonable to choose product operations rather than plus operations as the coherence of the whole document is related to the coherence of each clique, and any incoherent clique would have an extremely adverse impact to the entire document. For document pair $< D_1, D_2 >$, if $S_{D_1} > S_{D_2}$, we would say document $D_1$ is more coherent than $D_2$.

\section{Coherence Experiments}
We evaluate the proposed coherence modeling framework on a common evaluation task usually adopted in the existing literature: Sentence Ordering.\\
\textbf{Data.} We employ two corpora which are widely used in this task \cite{DBLP:conf/naacl/BarzilayL04,DBLP:conf/acl/ElsnerC08a,DBLP:journals/coling/BarzilayL08,DBLP:conf/acl/ElsnerC11a,DBLP:conf/emnlp/LouisN12}. The first is a collection of aviation accident reports written by officials from the National Transportation Safety Board and the second contains Associated Press articles from the North American News Corpus on the topic of earthquakes. 
%In our experiment, both two sets exhibit clear sentence structure with documents of comparable length and vocabulary size. 
The size of the word vocabulary $V$ for the experiments using accident corpus is  $4501$ and with approximately $11.5$ sentences per document on average. For the earthquake corpus, $|V| =3022$ with about $10.4$ sentences per document on average. 
%For each corpus, we use 100 source articles with 20 randomly generated permutations(2000 pairs) for training. We also held out 10 original documents and their permutations(200 pairs) from the training data for development purposes. 
Following the setup of \cite{DBLP:conf/emnlp/LiH14a}, 100 source articles are used for training, and 100 (accidents) and 99 (earthquakes) are used for testing. A maximum of 20 random permutations were generated for each test article to create the pairwise data. Positive cliques are directly obtained from the original training document and negative examples are created by random permutations of its sentences within the document. Moreover, like the method in \cite{DBLP:conf/sigir/SeverynM15}, we employ the word2vec tool to compute the word embeddings for sentence matrix construction.\\
%In accident corpus, test set consists of 100 source documents with of 1986 test pairs, and there are 99 original articles with 1956 pairs of data in earthquake test set.\\\\
%\textbf{Word embeddings\\}
%We adopt the embedding used in \cite{DBLP:conf/sigir/SeverynM15,DBLP:journals/corr/SeverynM16}. They run word2vec tool \cite{DBLP:conf/nips/MikolovSCCD13} on the English Wikipedia dump and the resulting model contains about 3.5 million word vectors dimensioned 50. For words that are not present in the word2vec model, their embeddings are randomly initialized with each component uniformly sampled from the uniform distribution $U[-0.25, 0.25]$. To reduce the size of the resulting vocabulary $V$ , they also replace all digits with 0. The final accident corpora ends up with 94\% of words initialized using wor2vec embeddings and the remaining 4\% words are initialized at random. Meanwhile, the earthquake corpora contains about 96\% words found in the word2vec model.
%
%\subsection{Evaluation Task}
%To enable direct comparison with the previous work, we evaluate performance of the proposed coherence model by the accuracy measured as the fraction of pairs in text set which can be correctly identified the original article  with higher score in the pair.
%\textbf{Baselines\\}
\textbf{Baselines.} To demonstrate that the CNN truly improves the coherence assessment performance in comparison with the state-of-the-art methods, we compare DCM with the following representative methods: Recursive \cite{DBLP:conf/emnlp/LiH14a}, Recurrent \cite{DBLP:conf/emnlp/LiH14a}, Entity Grid \cite{DBLP:journals/coling/BarzilayL08}, HMM \cite{DBLP:conf/emnlp/LouisN12}, HMM+Content \cite{DBLP:conf/emnlp/LouisN12}, Conference+Syntax \cite{DBLP:journals/coling/BarzilayL08}, and Graph \cite{DBLP:conf/acl/GuinaudeauS13}. 

In addition, to verify the effectiveness of the similarity building blocks in the deep learning architecture, we also study a configuration of the proposed model without the similarity component: DCM\_Nosim.

\subsection{Training and Hyper-parameters}
We train our deep learning architecture on a training set using stochastic gradient descent (SGD) and tune parameters of the network on a development set. The word embeddings matrix $\mathbf{W}$ has dimension 50 and the width of convolution filters is 4. %Additionally, we set the dimensionality of overlap feature matrix $d_o$ to 5 and randomly initialize the entries of the matrix $\mathbf{W}_o$ by sampling from the uniform distribution. 
There are 100 convolutional feature maps, such that each intermediate vector obtained in the sentence model has also dimension 100. 
%We apply two matrix $\mathbf{M} \in \mathbb{R}^{100\times100}$ to the model to compute the similarity between sentence vectors. These two similarity scores, together with three distributional vectors form a single joint representation which has 302 neurons, the same with hidden layer.
%We eliminate the need to tune the learning rate by using the Adadelta update rule \cite{DBLP:journals/corr/abs-1212-5701}.
Batch size is set to 500 examples and the network is trained for 20 epochs with early stopping.

\subsection{Results and Discussion}
Table~\ref{tab:result} reports the results of DCM and all the competing methods in the evaluation task. The experimental results are averaged with 10 random initializations. As we see, DCM achieves a much stronger performance than all the existing methods by a large margin, showing a significant improvement of about 5.3\% gain on average for the accident and earthquake corpora.

\begin{table}
\renewcommand\arraystretch{1}  
  \caption{Survey of the results with average accuracy in two corpora on the Sentence Ordering task.}
%\vspace{-0.5em}
  \label{tab:result}
  \begin{tabular}{lccc}
    \toprule
       \textbf{Model}            &  \textbf{Accident}  &\textbf{Earthquake}  &\textbf{Average}   \\ 
    \midrule
        \textbf{DCM}                         & \textbf{0.950}                           & \textbf{0.995}       &\textbf{ 0.973  }                 \\ 
     DCM\_Nosim            & 0.925                           & 0.986       & 0.956                   \\
	Recursive                   & 0.864                                    & {\color[HTML]{000000} 0.976}     & 0.920         \\ 
	Recurrent                   & 0.840                                    & 0.951      & 0.895                                \\ 
	Entity Grid                 & 0.904                                    & 0.872       & 0.888                                \\ 
	HMM                         & 0.822                                    & 0.938          &0.880                            \\ 
	HMM+Content                  & 0.742                                    & 0.953         & 0.848                          \\
	Conference+Syntax          & 0.765                                    & 0.888            & 0.827                           \\ 
	Graph                         & 0.846                                    & 0.635               & 0.741                         \\
\bottomrule
\end{tabular}
%\vspace{-1em}
\end{table}

Compared with HMM and Entity Grid, DCM requires no manual feature engineering anymore and can automatically learn better sentence representations using distributional word embeddings. 
%However, most of existing baselines and other traditional coherence methods require sophisticated feature selection processes and greatly rely on external feature extraction algorithm. Although many of them performs well in some specific articles, it is still hard to obtain suitable features for new documents from different genres and brings some challenges for considering accurate weight proportion between these manual features without sufficient professional experience. 
Further, the abstract sentence representations computed by DCM are more meaningful in exactly capturing the relevant semantic, logical and syntactic features in coherent context than all the competing models. %Besides, due to large width of the convolution filter, our neural network is able to learn advanced intermediate representations by capturing longer range dependencies in sentence, while other coherence architectures only operate on unigram or bigrams \cite{DBLP:conf/naacl/BarzilayL04} may be capable of extracting information from single word and short phrases but fails to detect some potential associations from words within long distance. 

Different from recursive neural network \cite{DBLP:conf/emnlp/LiH14a}, which asks for expensive preprocessing using syntactic parsers to construct syntactic trees and then builds the convolution on them, CNN does not require any NLP parsers for preprocessing or external semantic resources.

The superior performance of DCM over DCM\_Nosim demonstrates the necessity of the similarity computation in coherence assessment, while both recursive and recurrent neural networks \cite{DBLP:conf/emnlp/LiH14a} ignore this point and cannot achieve perfect results.

\section{Conclusion}
In this paper, we develop a deep coherence model, DCM, based on convolutional neural networks for text coherence assessment. The text coherence problem is investigated with a new perspective of learning sentence distributional representation and text coherence modeling simultaneously. In particular, DCM captures the interactions between sentences by computing the similarities of their distributional representations. Further, it can be easily trained in an end-to-end fashion. DCM is evaluated on a standard Sentence Ordering task. The experimental results demonstrate its effectiveness and promise in coherence assessment showing significant improvements over the state-of-the-art models by a wide margin.\\

%In terms of future work, an important direction lies in analyzing coherence for multiparagraph documents which are much longer and more challenging than the short documents we faced now. Instead of using a full-scale coherence model, a better approach may be cutting up these documents into more manageable chunks that can be handled as a unit. 

%The text above, the equation
%\begin{equation}
%(2n)!/\bigl(n!\,(n+1)!\bigr).
%\end{equation}
%The text below.
%The text above, the equation
%{\setlength\abovedisplayskip{1pt}
%\setlength\belowdisplayskip{1pt}
%\begin{equation}
%(2n)!/\bigl(n!\,(n+1)!\bigr).
%\end{equation}}
%The text below.

%\end{document}  % This is where a 'short' article might terminate

\begin{acks}
%  The authors would like to thank Dr. Yuhua Li for providing the
%  matlab code of  the \textit{BEPS} method. 
%
%  The authors would also like to thank the anonymous referees for
%  their valuable comments and helpful suggestions. The work is
%  supported by the \grantsponsor{GS501100001809}{National Natural
%    Science Foundation of
%    China}{http://dx.doi.org/10.13039/501100001809} under Grant
%  No.:~\grantnum{GS501100001809}{61273304}
%  and~\grantnum[http://www.nnsf.cn/youngscientsts]{GS501100001809}{Young
%    Scientsts' Support Program}.
We thank all reviewers for their valuable comments. This work was supported by National Natural Science Foundation of China (NSFC No. 61672456), the Fundamental Research Funds for the Central Universities (No. 2017QNA5008, 2017FZA5007), and Zhejiang Provincial Engineering Center on Media Data Cloud Processing and Analysis.

\end{acks}

% Bibliography
\bibliographystyle{ACM-Reference-Format}
%\vspace{-1em}
\bibliography{sigproccby}

%%% -*-BibTeX-*-
%%% Do NOT edit. File created by BibTeX with style
%%% ACM-Reference-Format-Journals [18-Jan-2012].

\begin{thebibliography}{00}

%%% ====================================================================
%%% NOTE TO THE USER: you can override these defaults by providing
%%% customized versions of any of these macros before the \bibliography
%%% command.  Each of them MUST provide its own final punctuation,
%%% except for \shownote{}, \showDOI{}, and \showURL{}.  The latter two
%%% do not use final punctuation, in order to avoid confusing it with
%%% the Web address.
%%%
%%% To suppress output of a particular field, define its macro to expand
%%% to an empty string, or better, \unskip, like this:
%%%
%%% \newcommand{\showDOI}[1]{\unskip}   % LaTeX syntax
%%%
%%% \def \showDOI #1{\unskip}           % plain TeX syntax
%%%
%%% ====================================================================

\ifx \showCODEN    \undefined \def \showCODEN     #1{\unskip}     \fi
\ifx \showDOI      \undefined \def \showDOI       #1{#1}\fi
\ifx \showISBNx    \undefined \def \showISBNx     #1{\unskip}     \fi
\ifx \showISBNxiii \undefined \def \showISBNxiii  #1{\unskip}     \fi
\ifx \showISSN     \undefined \def \showISSN      #1{\unskip}     \fi
\ifx \showLCCN     \undefined \def \showLCCN      #1{\unskip}     \fi
\ifx \shownote     \undefined \def \shownote      #1{#1}          \fi
\ifx \showarticletitle \undefined \def \showarticletitle #1{#1}   \fi
\ifx \showURL      \undefined \def \showURL       {\relax}        \fi
% The following commands are used for tagged output and should be
% invisible to TeX
\providecommand\bibfield[2]{#2}
\providecommand\bibinfo[2]{#2}
\providecommand\natexlab[1]{#1}
\providecommand\showeprint[2][]{arXiv:#2}

\bibitem[\protect\citeauthoryear{Barzilay and Lapata}{Barzilay and
  Lapata}{2008}]%
        {DBLP:journals/coling/BarzilayL08}
\bibfield{author}{\bibinfo{person}{Regina Barzilay} {and}
  \bibinfo{person}{Mirella Lapata}.} \bibinfo{year}{2008}\natexlab{}.
\newblock \showarticletitle{Modeling Local Coherence: An Entity-Based
  Approach}.
\newblock \bibinfo{journal}{{\em Computational Linguistics\/}}
  \bibinfo{volume}{34}, \bibinfo{number}{1} (\bibinfo{year}{2008}),
  \bibinfo{pages}{1--34}.
\newblock


\bibitem[\protect\citeauthoryear{Barzilay and Lee}{Barzilay and Lee}{2004}]%
        {DBLP:conf/naacl/BarzilayL04}
\bibfield{author}{\bibinfo{person}{Regina Barzilay} {and}
  \bibinfo{person}{Lillian Lee}.} \bibinfo{year}{2004}\natexlab{}.
\newblock \showarticletitle{Catching the Drift: Probabilistic Content Models,
  with Applications to Generation and Summarization}. In
  \bibinfo{booktitle}{{\em {HLT-NAACL}}}. \bibinfo{pages}{113--120}.
\newblock


\bibitem[\protect\citeauthoryear{Bengio, Simard, and Frasconi}{Bengio
  et~al\mbox{.}}{1994}]%
        {DBLP:journals/tnn/BengioSF94}
\bibfield{author}{\bibinfo{person}{Yoshua Bengio}, \bibinfo{person}{Patrice~Y.
  Simard}, {and} \bibinfo{person}{Paolo Frasconi}.}
  \bibinfo{year}{1994}\natexlab{}.
\newblock \showarticletitle{Learning long-term dependencies with gradient
  descent is difficult}.
\newblock \bibinfo{journal}{{\em {IEEE} Trans. Neural Networks\/}}
  \bibinfo{volume}{5}, \bibinfo{number}{2} (\bibinfo{year}{1994}),
  \bibinfo{pages}{157--166}.
\newblock


\bibitem[\protect\citeauthoryear{Bordes, Weston, and Usunier}{Bordes
  et~al\mbox{.}}{2014}]%
        {DBLP:conf/pkdd/BordesWU14}
\bibfield{author}{\bibinfo{person}{Antoine Bordes}, \bibinfo{person}{Jason
  Weston}, {and} \bibinfo{person}{Nicolas Usunier}.}
  \bibinfo{year}{2014}\natexlab{}.
\newblock \showarticletitle{Open Question Answering with Weakly Supervised
  Embedding Models}. In \bibinfo{booktitle}{{\em {ECML} {PKDD}}}.
  \bibinfo{pages}{165--180}.
\newblock


\bibitem[\protect\citeauthoryear{Elsner and Charniak}{Elsner and
  Charniak}{2008}]%
        {DBLP:conf/acl/ElsnerC08a}
\bibfield{author}{\bibinfo{person}{Micha Elsner} {and} \bibinfo{person}{Eugene
  Charniak}.} \bibinfo{year}{2008}\natexlab{}.
\newblock \showarticletitle{Coreference-inspired Coherence Modeling}. In
  \bibinfo{booktitle}{{\em {ACL}}}. \bibinfo{pages}{41--44}.
\newblock


\bibitem[\protect\citeauthoryear{Elsner and Charniak}{Elsner and
  Charniak}{2011}]%
        {DBLP:conf/acl/ElsnerC11a}
\bibfield{author}{\bibinfo{person}{Micha Elsner} {and} \bibinfo{person}{Eugene
  Charniak}.} \bibinfo{year}{2011}\natexlab{}.
\newblock \showarticletitle{Extending the Entity Grid with Entity-Specific
  Features}. In \bibinfo{booktitle}{{\em {ACL-HLT}}}.
  \bibinfo{pages}{125--129}.
\newblock


\bibitem[\protect\citeauthoryear{Fung and Ngai}{Fung and Ngai}{2006}]%
        {fung2006one}
\bibfield{author}{\bibinfo{person}{Pascale Fung} {and} \bibinfo{person}{Grace
  Ngai}.} \bibinfo{year}{2006}\natexlab{}.
\newblock \showarticletitle{One story, one flow: Hidden Markov Story Models for
  multilingual multidocument summarization}.
\newblock \bibinfo{journal}{{\em {TSLP}\/}} \bibinfo{volume}{3},
  \bibinfo{number}{2} (\bibinfo{year}{2006}), \bibinfo{pages}{1--16}.
\newblock


\bibitem[\protect\citeauthoryear{Grosz, Weinstein, and Joshi}{Grosz
  et~al\mbox{.}}{1995}]%
        {grosz1995centering}
\bibfield{author}{\bibinfo{person}{Barbara~J Grosz}, \bibinfo{person}{Scott
  Weinstein}, {and} \bibinfo{person}{Aravind~K Joshi}.}
  \bibinfo{year}{1995}\natexlab{}.
\newblock \showarticletitle{Centering: A framework for modeling the local
  coherence of discourse}.
\newblock \bibinfo{journal}{{\em Computational linguistics\/}}
  \bibinfo{volume}{21}, \bibinfo{number}{2} (\bibinfo{year}{1995}),
  \bibinfo{pages}{203--225}.
\newblock


\bibitem[\protect\citeauthoryear{Guinaudeau and Strube}{Guinaudeau and
  Strube}{2013}]%
        {DBLP:conf/acl/GuinaudeauS13}
\bibfield{author}{\bibinfo{person}{Camille Guinaudeau} {and}
  \bibinfo{person}{Michael Strube}.} \bibinfo{year}{2013}\natexlab{}.
\newblock \showarticletitle{Graph-based Local Coherence Modeling}. In
  \bibinfo{booktitle}{{\em {ACL}}}. \bibinfo{pages}{93--103}.
\newblock


\bibitem[\protect\citeauthoryear{Kim}{Kim}{2014}]%
        {DBLP:conf/emnlp/Kim14}
\bibfield{author}{\bibinfo{person}{Yoon Kim}.} \bibinfo{year}{2014}\natexlab{}.
\newblock \showarticletitle{Convolutional Neural Networks for Sentence
  Classification}. In \bibinfo{booktitle}{{\em {EMNLP}}}.
  \bibinfo{pages}{1746--1751}.
\newblock


\bibitem[\protect\citeauthoryear{Li and Hovy}{Li and Hovy}{2014}]%
        {DBLP:conf/emnlp/LiH14a}
\bibfield{author}{\bibinfo{person}{Jiwei Li} {and} \bibinfo{person}{Eduard~H.
  Hovy}.} \bibinfo{year}{2014}\natexlab{}.
\newblock \showarticletitle{A Model of Coherence Based on Distributed Sentence
  Representation}. In \bibinfo{booktitle}{{\em {EMNLP}}}.
  \bibinfo{pages}{2039--2048}.
\newblock


\bibitem[\protect\citeauthoryear{Louis and Nenkova}{Louis and Nenkova}{2012}]%
        {DBLP:conf/emnlp/LouisN12}
\bibfield{author}{\bibinfo{person}{Annie Louis} {and} \bibinfo{person}{Ani
  Nenkova}.} \bibinfo{year}{2012}\natexlab{}.
\newblock \showarticletitle{A Coherence Model Based on Syntactic Patterns}. In
  \bibinfo{booktitle}{{\em EMNLP-CoNLL}}. \bibinfo{pages}{1157--1168}.
\newblock


\bibitem[\protect\citeauthoryear{Mesnil, He, Deng, and Bengio}{Mesnil
  et~al\mbox{.}}{2013}]%
        {DBLP:conf/interspeech/MesnilHDB13}
\bibfield{author}{\bibinfo{person}{Gr{\'{e}}goire Mesnil},
  \bibinfo{person}{Xiaodong He}, \bibinfo{person}{Li Deng}, {and}
  \bibinfo{person}{Yoshua Bengio}.} \bibinfo{year}{2013}\natexlab{}.
\newblock \showarticletitle{Investigation of recurrent-neural-network
  architectures and learning methods for spoken language understanding}. In
  \bibinfo{booktitle}{{\em {INTERSPEECH}}}. \bibinfo{pages}{3771--3775}.
\newblock


\bibitem[\protect\citeauthoryear{Mikolov, Karafi{\'{a}}t, Burget,
  Cernock{\'{y}}, and Khudanpur}{Mikolov et~al\mbox{.}}{2010}]%
        {DBLP:conf/interspeech/MikolovKBCK10}
\bibfield{author}{\bibinfo{person}{Tomas Mikolov}, \bibinfo{person}{Martin
  Karafi{\'{a}}t}, \bibinfo{person}{Luk{\'{a}}s Burget}, \bibinfo{person}{Jan
  Cernock{\'{y}}}, {and} \bibinfo{person}{Sanjeev Khudanpur}.}
  \bibinfo{year}{2010}\natexlab{}.
\newblock \showarticletitle{Recurrent neural network based language model}. In
  \bibinfo{booktitle}{{\em {INTERSPEECH}}}. \bibinfo{pages}{1045--1048}.
\newblock


\bibitem[\protect\citeauthoryear{Mikolov, Sutskever, Chen, Corrado, and
  Dean}{Mikolov et~al\mbox{.}}{2013}]%
        {DBLP:conf/nips/MikolovSCCD13}
\bibfield{author}{\bibinfo{person}{Tomas Mikolov}, \bibinfo{person}{Ilya
  Sutskever}, \bibinfo{person}{Kai Chen}, \bibinfo{person}{Gregory~S. Corrado},
  {and} \bibinfo{person}{Jeffrey Dean}.} \bibinfo{year}{2013}\natexlab{}.
\newblock \showarticletitle{Distributed Representations of Words and Phrases
  and their Compositionality}. In \bibinfo{booktitle}{{\em {NIPS}}}.
  \bibinfo{pages}{3111--3119}.
\newblock


\bibitem[\protect\citeauthoryear{Poesio, Stevenson, Eugenio, and
  Hitzeman}{Poesio et~al\mbox{.}}{2004}]%
        {DBLP:journals/coling/PoesioSEH04}
\bibfield{author}{\bibinfo{person}{Massimo Poesio}, \bibinfo{person}{Rosemary
  Stevenson}, \bibinfo{person}{Barbara~Di Eugenio}, {and}
  \bibinfo{person}{Janet Hitzeman}.} \bibinfo{year}{2004}\natexlab{}.
\newblock \showarticletitle{Centering: {A} Parametric Theory and Its
  Instantiations}.
\newblock \bibinfo{journal}{{\em Computational Linguistics\/}}
  \bibinfo{volume}{30}, \bibinfo{number}{3} (\bibinfo{year}{2004}),
  \bibinfo{pages}{309--363}.
\newblock


\bibitem[\protect\citeauthoryear{Severyn and Moschitti}{Severyn and
  Moschitti}{2015}]%
        {DBLP:conf/sigir/SeverynM15}
\bibfield{author}{\bibinfo{person}{Aliaksei Severyn} {and}
  \bibinfo{person}{Alessandro Moschitti}.} \bibinfo{year}{2015}\natexlab{}.
\newblock \showarticletitle{Learning to Rank Short Text Pairs with
  Convolutional Deep Neural Networks}. In \bibinfo{booktitle}{{\em {ACM}
  {SIGIR}}}. \bibinfo{pages}{373--382}.
\newblock


\bibitem[\protect\citeauthoryear{Severyn and Moschitti}{Severyn and
  Moschitti}{2016}]%
        {DBLP:journals/corr/SeverynM16}
\bibfield{author}{\bibinfo{person}{Aliaksei Severyn} {and}
  \bibinfo{person}{Alessandro Moschitti}.} \bibinfo{year}{2016}\natexlab{}.
\newblock \showarticletitle{Modeling Relational Information in Question-Answer
  Pairs with Convolutional Neural Networks}.
\newblock \bibinfo{journal}{{\em CoRR\/}}  \bibinfo{volume}{abs/1604.01178}
  (\bibinfo{year}{2016}).
\newblock


\end{thebibliography}

\end{document}